\documentclass{article}
\usepackage{ijcai18}
 
\usepackage{times}
\usepackage[svgnames,x11names]{xcolor}
\usepackage{soul}
\usepackage[utf8]{inputenc}
\usepackage[small]{caption}

\usepackage{url}
\usepackage{multirow}
\usepackage{epsfig}
\usepackage{amsmath}
\usepackage{subcaption}
\usepackage{graphicx}
\usepackage{bm}
\usepackage{color,xcolor,framed}
\usepackage{multirow}
\newcommand*{\affaddr}[1]{#1} 
\newcommand*{\affmark}[1][*]{\textsuperscript{#1}}
\newcommand*{\email}[1]{#1}

\newcommand{\cev}[1]{\reflectbox{\ensuremath{\vec{\reflectbox{\ensuremath{#1}}}}}}

\newcommand{\mycolorbox}[2]{\setlength{\fboxsep}{2pt}\colorbox{#1}{\strut #2}}

\title{A Hierarchical End-to-End Model for Jointly Improving Text Summarization\\ and Sentiment Classification}

\author{Shuming Ma\affmark[1], Xu Sun\affmark[1], Junyang Lin\affmark[2], Xuancheng Ren\affmark[1]\\
\affaddr{\affmark[1]MOE Key Lab of Computational Linguistics, School of EECS, Peking University}\\
\affaddr{\affmark[2]School of Foreign Languages, Peking University}\\
\email{\{shumingma, xusun, linjunyang, renxc\}@pku.edu.cn}\\
}

\begin{document}

\maketitle

\begin{abstract}
Text summarization and sentiment classification both aim to capture the main ideas of the text but at different levels. Text summarization is to describe the text within a few sentences, while sentiment classification can be regarded as a special type of summarization which ``summarizes'' the text into a even more abstract fashion, i.e., a sentiment class. Based on this idea, we propose a hierarchical end-to-end model for joint learning of text summarization and sentiment classification, where the sentiment classification label is treated as the further ``summarization'' of the text summarization output. Hence, the sentiment classification layer is put upon the text summarization layer, and a hierarchical structure is derived. Experimental results on Amazon online reviews datasets show that our model achieves better performance than the strong baseline systems on both abstractive summarization and sentiment classification.
\end{abstract}

\section{Introduction}

Text summarization and sentiment classification are two important tasks in natural language processing. Text summarization aims at generating a summary with the major points of the original text. Compared with extractive summarization, which selects a subset of existing words in the original text to form the summary, abstractive summarization builds an internal semantic representation and then uses natural language generation techniques to create a summary that is closer to what a human might express. In this work, we mainly focus on the abstractive text summarization. Sentiment classification is to assign a sentiment label to determine the attitude or the opinion inside the text. It is also known as opinion mining, deriving the opinion or the attitude of a speaker. Both text summarization and sentiment classification aim at mining the main ideas of the text. Text summarization describes the text with words and sentences in a more specific way, while sentiment classification summarizes the text with labels in a more abstractive way.

Most of the existing models are built for either summarization or classification.
For abstractive text summarization, the most popular model is the sequence-to-sequence model~\cite{seq2seq,abs}, where generating a short summary for the long source text can be regarded as a mapping between a long sequence and a short sequence. The model consists of an encoder and a decoder. The encoder encodes the original text into a latent representation, and the decoder generates the summary. Some recent abstractive summarization models are the variants of the sequence-to-sequence model~\cite{ras,See2017}.
For sentiment classification, most of the recent work uses the neural network architecture~\cite{Kim2014,Tang2015}, such as LSTM or CNN, to generate a text embedding, and use a multi-layer perceptron (MLP) to predict the label from the embedding.

Some previous work~\cite{hole2013,mane2015} proposes the models to produce both the summaries and the sentiment labels. However, these models train the summarization part and the sentiment classification part independently, and require rich, hand-craft features. There are also some work about the sentiment summarization~\cite{Titov2008,Lerman2009}, which aim at extracting the sentences  with a certain sentiment class from the original texts. These work only focuses on the summarization, and does not improve the sentiment classification.

In this work, we explore a first step towards improving both text summarization and sentiment classification within an end-to-end framework.
We propose a hierarchical end-to-end model, which consists of a summarization layer and a sentiment classification layer.
The summarization layer compresses the original text into short sentences, and the sentiment classification layer further ``summarizes'' the texts into a sentiment class.
The hierarchical structure establishes a close bond between text summarization and sentiment classification, so that the two tasks can improve each other.
After compressing the texts with summarization, it will be easier for the sentiment classifier to predict the sentiment labels of the shorter text. Besides, text summarization can point out the important and informative words, and remove the redundant and misleading information that is harmful to predict the sentiment.
The sentiment classification can provide a more significant supervision signal for text summarization, and guides the summarization component to capture the sentiment tendency of the original text, which can improve the coherence between the short text and the original text.

We evaluate our proposed model on Amazon online reviews datasets. Experimental results show that our model achieves better performance than the strong baseline systems on both summarization and sentiment classification. 

The contributions of this paper are listed as follows:
\begin{itemize}
\item We treat the sentiment classification as a special type of summarization, and perform sentiment classification and text summarization using a unified model.
\item We propose a multi-view attention to obtain different representation of the texts for summarization and sentiment classification.
\item Experimental results shows that our model outperforms the strong baselines that train the summarization and sentiment classification separately.
\end{itemize}

\section{Proposed Model}

In this section, we introduce our proposed model in details. In Section~\ref{problem}, we give the problem formulation. We explain the overview of our proposed model in Section~\ref{overview}. Then, we introduce each components of the model from Section~\ref{encoder} to Section~\ref{sec_classifier}. Finally, Section~\ref{training} gives the overall loss function and the training methods.

\subsection{Problem Formulation}\label{problem}

Given an online reviews dataset that consists of $N$ data samples, the $i$-th data sample ($x^{i}$, $y^{i}$, $l^{i}$) contains an original text $x^{i}$, a summary $y^{i}$, and a sentiment label $l^{i}$. Both the original content $x^{i}$ and the summary $y^{i}$ are sequences of words:
\begin{equation*}
x^{i}=\{x^{i}_{1},x^{i}_{2},...,x^{i}_{L_{i}}\}
\end{equation*}
\begin{equation*}
y^{i}=\{y^{i}_{1},y^{i}_{2},...,y^{i}_{M_{i}}\}
\end{equation*}
where $L_{i}$ and $M_{i}$ denote the number of words in the sequences $x^{i}$ and $y^{i}$, respectively. The label $l^{i} \in \{1,2,...,K\}$ denotes the sentiment attitude of the original content $x^{i}$, from the lowest rating $1$ to the highest rating $K$.

The model is applied to learn the mapping from the source text to the target summary and the sentiment label. For the purpose of simplicity, $(\bm{x}, \bm{y}, l)$ is used to denote each data
pair in the rest of this section, where $\bm{x}$ is the word sequence of an original text, $\bm{y}$ is the word sequence of the corresponding summary, and $l$ is the corresponding sentiment label.

\begin{figure}[tb]
	\centering
	\includegraphics[width=1.0\linewidth]{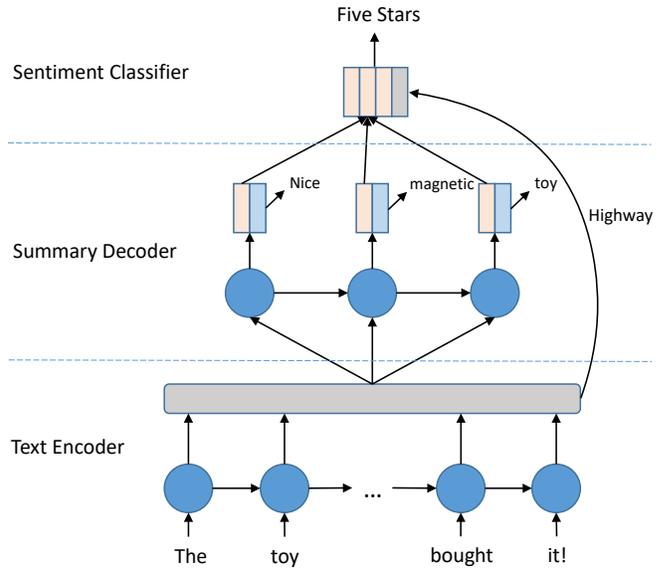}
	\caption{The overview of our model.}\label{model_fig}
	\vspace{-0.05in}
\end{figure}

\subsection{Model Overview}\label{overview}

Figure~\ref{model_fig} shows the architecture of our model. Our model consists of three components, which are the text encoder, the summary decoder, and the sentiment classifier. The text encoder compresses the original text $\bm{x}$ into the context memory $\bm{h}$ with a bi-directional LSTM. The summary decoder is a uni-directional LSTM, which then generates a summary vector $v^{(c)}$ and a sentiment vector $v^{(t)}$ sequentially with the attention mechanism by querying the context memory $\bm{h}$. The summary vectors $v^{(c)}$ are used to generate the summary with a word generator. The sentiment vectors $v^{(t)}$ of all time steps are collected and then fed into the sentiment classifier to predict the sentiment label. In order to capture the context information of the original text, we use the highway mechanism to feed the the context memory $\bm{h}$ as part of the input of the classifier. Therefore, the classifier predicts the label according to the sentiment vectors of the summary decoder and the context memory of the text encoder.

\subsection{Text Encoder}\label{encoder}

The goal of the source text encoder is to provide a series of dense representation of the original text for the decoder and the classifier. In our model, the original text encoder is a bi-directional Long Short-term Memory Network (BiLSTM), which produces the context memory $\bm{h}=\{h_1,h_2,...,h_L\}$ from the source text $\bm{x}$:
\begin{equation}\label{flstm}
\vec{h}_t = \vec{f}(x_t,\vec{h}_{t-1})
\end{equation}
\begin{equation}\label{blstm}
\cev{h}_t = \cev{f}(x_t,\cev{h}_{t+1})
\end{equation}
\begin{equation}\label{clstm}
h_t = \vec{h}_t+\cev{h}_t
\end{equation}
where $\vec{f}$ and $\cev{f}$ are the forward and the backward functions of LSTM for one time step, $\vec{h}_t$ and $\cev{h}_t$ are the forward and the backward hidden outputs respectively, $x_t$ is the input at the $t$-th time step, and $L$ is the number of words in sequence $\bm{x}$.

Although convolutional neural network (CNN) is also an alternative choice for the encoder, BiLSTM is more popular for the sequence-to-sequence learning of text generation tasks including abstractive text summarization. Besides, according to our experiments, BiLSTM achieves better performance in sentiment classification on our benchmark datasets. We give the details of the comparison of CNN and BiLSTM in Section~\ref{experiment}.

\subsection{Summary Decoder with Multi-View Attention}\label{decoder}

The goal of the summary decoder is to generate a series of summary words, and provides the summary information for the sentiment classifier. In our model, the summary decoder consists of a uni-directional LSTM, a multi-view attention mechanism, and a word generator. The LSTM first generates the hidden output $s_t$ conditioned on the historical information of the generated summary:
\begin{equation}\label{flstm}
s_t = f(y_{t-1},s_{t-1})
\end{equation}
where $f$ is the function of LSTM for one time step, and $y_{t-1}$ is the last generated words at $t$-th time step.

Given the hidden output $s_t$, we implement a multi-view attention mechanism to retrieval the summary information and the sentiment information from the context memory $\bm{h}$ of the original text. The motivation of the multi-view attention is that the model should focus on different part of the original text for summarization and classification. For summarization, the attention mechanism should focus on the informative words that describe the main points best. For sentiment classification, the attention mechanism should focus on the words that contains the most sentimental tendency, such as ``great'', ``bad'', and so on. In implementation, the multi-view attention generates a summary vector $v^{(c)}$ for summarization:
\begin{equation}\label{attention1}
v_{t}^{(c)}=\sum_{i=1}^{N}{\alpha_{ti}h_{i}}
\end{equation}
\begin{equation}\label{attention2}
\alpha_{ti}=\frac{e^{g(s_{t},h_{i})}}{\sum_{j=1}^{N}{e^{g(s_{t},h_{j})}}}
\end{equation}
\begin{equation}\label{attention3}
g(s_{t},h_{i})=\tanh{(s^{T}_{t}W_{t}h_{i})}
\end{equation}
where $W_{t}$ is a trainable parameter matrix. Similar to the summary vector, the sentiment vector $v^{(t)}$ is also generated with the attention mechanism following Equation~\ref{attention1}, \ref{attention2}, and \ref{attention3}, but has different trainable parameters. The multi-view attention can be regarded as two independent global attentions to learn to focus more on the summary aspect or the sentiment aspect.

Given the summary vector $v_{t}^{(c)}$, the word generator is used to compute the probability distribution of the output words at $t$-th time step:
\begin{equation}\label{generator}
p(y_t|x)=softmax{(W_{g}v_{t}^{(c)}+b_{g})}
\end{equation}
where $W_{g}$ and $b_{g}$ are parameters of the generator. The word with the highest probability is emitted as $t$-th word of the generated summary.

\subsection{Summary-Aware Sentiment Classifier}\label{sec_classifier}

After decoding the words until the end of the summary, the model collects the sentiment vectors of all time step:
\begin{equation}
\bm{v^{(t)}}=[v^{(t)}_{1},v^{(t)}_{2},...,v^{(t)}_{M}]
\end{equation}
Then, we concatenate the summary sentiment vectors $\bm{v^{(t)}}$ and the original text representation $\bm{h}$, and perform a max-pooling operation to obtain a sentiment context vector $\bm{r}$, which we denote as a highway operation in Figure~\ref{model_fig}:
\begin{equation}
\begin{split}
\bm{r}&=\max(\bm{v^{(t)}}\oplus\bm{h}) \\
      &=\max([v^{(t)}_{1},v^{(t)}_{2},...,v^{(t)}_{M},h_{1},h_{2},...,h_{L}])
\end{split}
\end{equation}
where $\oplus$ denotes the operation of concatenation along the first dimension, $M$ is the number of words in the summary, and $L$ is the number of words in the original text. The sentiment context vector is then fed into the classifier to compute the probability distribution of the sentiment label $p(l|x)$.
The classifier is a two-layer feed-forward network with RELU as the activation function. The label with the highest probability is the predicted sentiment label.

\subsection{Overall Loss Function and Training}\label{training}

The loss function consists of two parts, which are the cross entropy loss of summarization and that of sentiment classification:
\begin{equation}
L_{s}=-\sum_t{{y_t}\log{p(y_t|x)}}
\end{equation}
\begin{equation}
L_{c}=-{l}\log{p(l|x)}
\end{equation}
where ${y_t}$ and ${l}$ are the ground truth of words and labels, and $p(y_t|x)$ and $p(l|x)$ are the probability distribution of words and labels, computed by Equation~\ref{generator}. We jointly minimize the two losses with Adam~\cite{KingmaBa2014} optimizer:
\begin{equation}
L=L_{s} + \lambda L_{c}
\end{equation}
where $\lambda$ is a hyper-parameter to balance two losses. We set $\lambda=0.5$ in this work.

\section{Experiments}\label{experiment}

In this section, we evaluate our model on the Amazon online review dataset, which contains the online reviews, summaries, and sentiment labels. We first introduce the datasets, evaluation metrics, and experimental details. Then, we compare our model with several popular baseline
systems. Finally, we provide the analysis and the discussion of our model.

\subsection{Datasets}

\noindent\textbf{Amazon SNAP Review Dataset (SNAP):} This dataset is part of \textbf{S}tanford \textbf{N}etwork \textbf{A}nalysis \textbf{P}roject (SNAP)\footnote{http://snap.stanford.edu/data/web-Amazon.html}, and is provided by He and McAuley~\shortcite{snap}. The dataset consists of reviews from Amazon, and contains product reviews and metadata from Amazon, including 142.8 million reviews spanning May 1996 - July 2014. It includes review content, product, user information, ratings, and summaries. We pair each review content with the corresponding summary and sentiment label. We select three domains of product reviews to construct three benchmark datasets, which are \textbf{Toys \& Games}, \textbf{Sports \& Outdoors}, and \textbf{Movie \& TV}. We select the first 1,000 samples of each dataset as the validation set, the following 1,000 samples as the test set, and the rest as the training set.

\subsection{Evaluation Metric}

For abstractive summarization, our evaluation metric is ROUGE score~\cite{rouge}, which is popular for summarization evaluation. The metrics compare an automatically produced summary with the reference summaries, by computing overlapping lexical units, including unigram, bigram, trigram, and longest common subsequence (LCS). Following previous work~\cite{abs,lcsts}, we use ROUGE-1 (unigram), ROUGE-2 (bi-gram) and ROUGE-L (LCS) as the evaluation metrics in the reported experimental results.

For sentiment classification, the evaluation metric is per-label accuracy. We evaluate the accuracy of both five-class sentiment, of which the sentiment is classified into 5 class, and two-class sentiment, of which the sentiment is either positive or negative.

\subsection{Experimental Details}

\subsubsection{Model Parameters}\label{model_param}

The vocabularies are extracted from the training sets, and the source contents and the summaries share the same vocabularies. We tune the hyper-parameters based on the performance on the validation sets.

We limit the vocabulary to 50,000 most frequent words appearing in the training set. We set the word embedding and the hidden size to 256, 512, and 512 for Toys, Sports, and Movies datasets, respectively. The word embedding is random initialized and learned from scratch. The encoder is a single-layer bidirectional LSTM, the decoder is a single-layer unidirectional LSTM, and the classifier is a two layer feed-forward network with a 512 hidden dimension. The batch size is 64, and we use dropout with probability $p = 0.2,\ 0.05,\ 0.0$ for Toys, Sports, and Movies datasets, respectively.

\subsubsection{Model Training}\label{model_train}

We use the Adam~\cite{KingmaBa2014} optimization method to train the model. For the hyper-parameters of Adam optimizer, we set the learning rate $\alpha = 0.0003$, two momentum parameters $\beta_{1} = 0.9$ and $\beta_{2} = 0.999$
respectively, and $\epsilon = 1 \times 10^{-8}$. Following Sutskever et al.~\shortcite{seq2seq}, we train the model for a total of 10 epochs, and start to halve the learning rate every half epoch after 5 epochs. We clip the gradients~\cite{gradientclip} to the maximum norm of 10.0.

\subsection{Baselines}

For abstractive summarization, our baseline model is the sequence-to-sequence model for abstractive summarization, following the previous work~\cite{lcsts}. We denote the sequence-to-sequence model without the attention mechanism as \textbf{S2S}, and that with the attention mechanism as \textbf{S2S-att}. 

For text classification, we compare our model with two baseline models: \textbf{BiLSTM} and \textbf{CNN}. For the two baseline models, the BiLSTM model uses a bidirectional LSTM with the dimension of 256 in each direction, and uses max pooling across all LSTM hidden states to get the sentence embedding vector, and then uses an MLP output layer with 512 hidden states to output the classification result. The CNN model uses the same scheme, but substitutes BiLSTM with 1 layer of convolutional network. During training we use 0.2 dropout on the MLP. We use Adam as the optimizer, with a learning rate of 0.001, and a batch size of 64. For BiLSTM, we also clip the norm of gradients to be 5.0. We searched hyper-parameters in a wide range and find the aforementioned set of hyperparameters yield the highest accuracy.

The above baseline models only exploit part of the annotated data (either summaries or sentiment labels). For fairer comparison, we also implement a joint model of S2S-att and BiLSTM (\textbf{S2S-att + BiLSTM}), and both the annotated labels of summaries and sentiments are used to train this baseline model. We compare our model with this model, in order to analyze the improvements of our model given exactly the same annotated data. In this baseline model, S2S-att and BiLSTM share the same encoder, and the S2S-att produces the summary with a LSTM decoder, while the BiLSTM predicts the sentiment label with a MLP. We tune the hyper-parameter on the validation set. We set the word embedding and the hidden size to 256, 512, and 512. The batch size is 64, and the dropout rate is $p=0.15,\ 0.05,\ 0.1$ for Toys, Sports, and Movies datasets, respectively.

\begin{table}[tb]
	\centering
	\begin{tabular}{l c c c }
		\hline
		\multicolumn{1}{l}{\textbf{Toys \& Games}} &
		\multicolumn{1}{c}{\textbf{RG-1}} & 
		\multicolumn{1}{c}{\textbf{RG-2}} &  
		\multicolumn{1}{c}{\textbf{RG-L}}   \\ \hline
		S2S~\cite{lcsts} & 14.05 & 2.47 & 15.75 \\
		S2S-att~\cite{lcsts} & 16.23 & 4.27 & 16.01 \\
		S2S-att + BiLSTM & 16.32 & 4.43 & 16.27 \\ \hline
		\textbf{HSSC (this work)} & \textbf{18.44} & \textbf{5.00} & \textbf{17.69} \\ 
        \hline
        \\
        \hline
		\multicolumn{1}{l}{\textbf{Sports \& Outdoors}} &
		\multicolumn{1}{c}{\textbf{RG-1}} & 
		\multicolumn{1}{c}{\textbf{RG-2}} &  
		\multicolumn{1}{c}{\textbf{RG-L}}   \\ \hline
		S2S~\cite{lcsts} & 13.38 & 2.59 & 13.18 \\
		S2S-att~\cite{lcsts} & 15.70 & 3.61 & 15.53 \\
		S2S-att + BiLSTM & 15.75 & 3.64 & 15.68 \\ \hline
		\textbf{HSSC (this work)} & \textbf{17.85} & \textbf{4.77} & \textbf{17.59} \\ 
        \hline
        \\
        \hline
		\multicolumn{1}{l}{\textbf{Movie \& TV}} &
		\multicolumn{1}{c}{\textbf{RG-1}} & 
		\multicolumn{1}{c}{\textbf{RG-2}} &  
		\multicolumn{1}{c}{\textbf{RG-L}}   \\ \hline
		S2S~\cite{lcsts} & 10.98 & 2.34 & 10.77 \\
		S2S-att~\cite{lcsts} & 12.17 & 3.08 & 11.77 \\
		S2S-att + BiLSTM & 12.33 & 3.22 & 11.92 \\ \hline
		\textbf{HSSC (this work)} & \textbf{14.52} & \textbf{4.84} & \textbf{13.42} \\ 
        \hline
		
	\end{tabular}
	\caption{Comparison between our model and the sequence-to-sequence baseline for abstractive summarization on the Amazon SNAP test sets. The test sets include three domains: Toys \& Gamse, Sports \& Outdoors, and Movie \& TV. RG-1, RG-2, and RG-L denote ROUGE-1, ROUGE-2, and ROUGE-L, respectively.}\label{tab_ats}
    \vspace{-0.15in}
\end{table}

\begin{table}[tb]
	\centering
	\begin{tabular}{l c c}
		\hline
		\multicolumn{1}{l}{\textbf{Toys \& Games}} &
		\multicolumn{1}{c}{\textbf{5-class}} & 
		\multicolumn{1}{c}{\textbf{2-class}}  \\ \hline
		CNN & 70.5 & 90.2 \\
		BiLSTM & 70.7 & 90.9  \\
		BiLSTM + S2S-att & 70.9 & 90.9  \\ \hline
		\textbf{HSSC (this work)} & \textbf{71.9} & \textbf{91.8}  \\ 
        \hline
        \\
        \hline
		\multicolumn{1}{l}{\textbf{Sports \& Outdoors}} &
		\multicolumn{1}{c}{\textbf{5-class}} & 
		\multicolumn{1}{c}{\textbf{2-class}}  \\ \hline
		CNN & 72.0 & 91.5 \\
		BiLSTM & 71.9 & 91.6  \\
		BiLSTM + S2S-att & 72.1 & 91.9  \\ \hline
		\textbf{HSSC (this work)} & \textbf{73.2} & \textbf{92.1}  \\ 
        \hline
        \\
        \hline
		\multicolumn{1}{l}{\textbf{Movie \& TV}} &
		\multicolumn{1}{c}{\textbf{5-class}} & 
		\multicolumn{1}{c}{\textbf{2-class}}  \\ \hline
		CNN & 66.9 & 86.0 \\
		BiLSTM & 67.8 & 86.2  \\
		BiLSTM + S2S-att & 68.0 & 86.6  \\ \hline
		\textbf{HSSC (this work)} & \textbf{68.9} & \textbf{88.4}  \\ 
        \hline
		
	\end{tabular}
	\caption{Comparison between our model and the sequence-to-sequence baselines for sentiment classification on the Amazon SNAP test sets. The test sets include three domains: Toys \& Games, Sports \& Outdoors, and Movie \& TV. 5-class and 2-class denote the accuracy of five-class sentiment and two-class sentiment, respectively.}\label{tab_tc}
    \vspace{-0.15in}
\end{table}

\subsection{Results}

We denote our \textbf{H}ierarchical \textbf{S}ummarization and \textbf{S}entiment \textbf{C}lassification model as \textbf{HSSC}.

\subsubsection{Abstractive Summarization}

First, we compare our model with the sequence-to-sequence baseline on the Amazon SNAP test sets.
We report the ROUGE F1 score of our model and the baseline models on the test sets.
As shown in Table~\ref{tab_ats}, our HSSC model has a large margin over both S2S and S2s-att models on all of the three test sets, which shows that the supervision of the sentiment labels improves the representation of the original text.
Moreover, given exactly the same annotated data (summary + sentiment label), our HSSC model still has an improvement over the S2S-att + BiLSTM baseline, which indicates that HSSC learns a better representation for summarization.
Overall, HSSC achieves the best performance in terms of ROUGE-1, ROUGE-2, and ROUGE-L over the three baseline models on the three test sets.

The summarization task on the online review texts is much more difficult and complicate, so the ROUGE scores on the SNAP dataset are lower than other summarization datasets, such as DUC. The documents in DUC datasets are originally from news website, so the texts are formal, and the summaries in DUC are manually selected and well written. The SNAP dataset is constructed with the reviews on the amazon, and both the original reviews and the corresponding summaries are informal and full of noise.

\subsubsection{Sentiment Classification}

We compare our model with two popular sentiment classification methods, which are CNN and BiLSTM, on the Amazon SNAP test sets.
We report the accuracy of five-grained sentiment and two-class sentiment on the test sets.
As shown in Table~\ref{tab_tc}, BiLSTM has a slightly improvement over the CNN baseline, showing that BiLSTM has a better performance to represent the texts on these datasets. Therefore, we select BiLSTM as the encoder of our model.
HSSC obtains a better performance over the two widely-used baseline models on all of the test sets, mainly because of the benefit of more labeled data and better representation.
What's more, HSSC outperforms the S2S-att + BiLSTM baseline, showing that the information from summary decoder helps to predict the sentiment labels.
Overall, HSSC achieves the best performance in terms of 5-class accuracy and 2-class accuracy over the three baseline models on the three test sets.

We have conducted significance tests based on t-test. The significance tests suggest that HSSC has a very significant improvement over all of the baselines, with p $\leq$ 0.001 in all of ROUGE metrics for summarization in three benchmark datasets, p $\leq$ 0.005 for sentiment classification in both Toys \& Games and Movies \& TV datasets, and p $\leq$ 0.01 for sentiment classification in the Sports \& Outdoors datasets.

\subsection{Ablation Study}

In order to analyze the effect of each components, We remove the components of multi-view and highway in order, and evaluate the performance of the rest model. We first remove the multi-view attention. As shown in Table~\ref{tab_ablation}, the model without multi-view attention has a drop of performance on both 5-class accuracy and ROUGE-L. It can be concluded that the multi-view attention improves the performance of both abstractive summarization and sentiment classification. We further remove the highway part, and find the highway component benefits not only the sentiment classification, bot also the abstractive summarization. The benefit mainly comes from the fact that the gradient of the sentiment classifier can be  directly propagated to the encoder, so that it learns a better representation of the original text for both classification and summarization.

\begin{table}[tb]
	\centering
	\begin{tabular}{l c c}
		\hline
		\multicolumn{1}{l}{\textbf{Toys \& Games}} &
		\multicolumn{1}{c}{\textbf{5-class}} & 
		\multicolumn{1}{c}{\textbf{RG-L}}  \\ \hline
        w/o Multi-View & 70.9 & 16.47  \\ 
        w/o Highway & 70.1 & 16.06  \\ 
		HSSC (Full Model) & 71.9 & 17.69  \\ 
        \hline
        \\
        \hline
		\multicolumn{1}{l}{\textbf{Sports \& Outdoors}} &
		\multicolumn{1}{c}{\textbf{5-class}} & 
		\multicolumn{1}{c}{\textbf{RG-L}}  \\ \hline
        w/o Multi-View & 72.0 & 16.36  \\ 
        w/o Highway & 71.5 & 15.73  \\ 
		HSSC (Full Model) & 73.2 & 17.59  \\ 
        \hline
        \\
        \hline
		\multicolumn{1}{l}{\textbf{Movie \& TV}} &
		\multicolumn{1}{c}{\textbf{5-class}} & 
		\multicolumn{1}{c}{\textbf{RG-L}}  \\ \hline
        w/o Multi-View & 68.1 & 12.34  \\ 
        w/o Highway & 67.7 & 12.01  \\ 
		HSSC (Full Model) & 68.9 & 13.42  \\ 
        \hline
	\end{tabular}
	\caption{Ablation study. 5-class denotes the accuracy of five-grained sentiment, and RG-L denotes ROUGE-L for summarization.}\label{tab_ablation}
    \vspace{-0.15in}
\end{table}

\subsection{Visualization of Multi-View Attention}

As shown in Table~\ref{visualization}, we present the heatmap of the attention scores of three examples. The multi-view attention allows the model to represent the text from the sentiment view and from the summary view. In order to analyze whether the multi-view attention captures the sentiment information and the summary information, we give the heatmap of the sentiment-view attention and the summary-view attention, respectively. We take the average of the attention scores in the decoder outputs at all time steps, and mark the high scores with deep color and the low scores with light color. From the table, we conclude that the sentiment-view attention focuses more on the sentimental words, e.g. ``\emph{best}'', ``\emph{powerful}'', ``\emph{great}'', ``\emph{fun}'', and ``\emph{comfortable}''. The summary-view attention concentrates on the informative words that best describes the opinion of the authors, e.g. ``\emph{i think that this is one of the best movie}'', and ``\emph{a great book, very fun}''. Moreover, the sentiment-view attention focuses more on the individual words, while the summary-view pays more attention on the word sequences. Besides, the sentiment-view attention and the summary-view attention share the focus on the informative words, showing the benefit from the multi-view attention.

\begin{table*}[tb]
	\small
	\centering
	\begin{tabular}{ l p{15.5cm}@{~} }
\hline
		(1) & \mycolorbox{MistyRose1}{i saw this movie}\mycolorbox{LightPink1}{11 times in the theater and}\mycolorbox{MistyRose1}{i think}\mycolorbox{LightPink1}{that this}\mycolorbox{IndianRed1}{is one of the best}\mycolorbox{MistyRose1}{movies ever made and the}\mycolorbox{IndianRed1}{best movie} \mycolorbox{MistyRose1}{made about}\mycolorbox{LightPink1}{christ}\mycolorbox{MistyRose1}{and his}\mycolorbox{LightPink1}{passion}\mycolorbox{MistyRose1}{. god bless all those}\mycolorbox{LightPink1}{responsible for the creation}of\mycolorbox{IndianRed1}{this powerful film}. \\ \\
        (2) & \mycolorbox{MistyRose1}{my daughter  ,  who is now 8 years old  ,  received this as a }\mycolorbox{LightPink1}{christmas gift} when she was 2 . it has been ready \mycolorbox{MistyRose1}{many times ,} \mycolorbox{MistyRose1}{and since been passed along}to my son who is now 4 . my children \mycolorbox{IndianRed1}{enjoy}the tactile quality of the monkeys faces .\mycolorbox{MistyRose1}{it is helpful learning counting when there is something they can feel}. i have\mycolorbox{IndianRed1}{always enjoyed}reading the sing song story . it does not take long to read , and after all these years i \mycolorbox{LightPink1}{pretty much} have it memorized . \mycolorbox{IndianRed1}{a great}\mycolorbox{LightPink1}{ book , }\mycolorbox{IndianRed1}{very fun }. \\ \\
        (3) & this mattress is \mycolorbox{IndianRed1}{too narrow}\mycolorbox{MistyRose1}{to be}\mycolorbox{IndianRed1}{comfortable}. you \mycolorbox{LightPink1}{fit on it fine but} because of the air ,\mycolorbox{MistyRose1}{i found that}it was a balancing act to switch positions . i \mycolorbox{LightPink1}{tried more and less} air to no effect . i think if you sleep on your\mycolorbox{MistyRose1}{back and stay in that position}it would be\mycolorbox{IndianRed1}{fine}but\mycolorbox{IndianRed1}{unfortunately}that is not how i sleep . the strong vinyl\mycolorbox{LightPink1}{smell}\mycolorbox{MistyRose1}{does go away after} airing out though . \\
%
		\hline
        \multicolumn{2}{c}{\multirow{2}{*}{(a) Sentiment view of the original text.}}\\ \\
        \hline
		(1) & \mycolorbox{MistyRose1}{i saw this movie 11 times in the theater and}\mycolorbox{IndianRed1}{i think that this is one of the best movies}\mycolorbox{LightPink1}{ever made and the}\mycolorbox{IndianRed1}{best movie made} \mycolorbox{MistyRose1}{about christ and his passion .}god bless all those responsible for the creation of this powerful film . \\ \\ 
        (2) & \mycolorbox{MistyRose1}{my daughter  ,  who is now 8 years old  ,  received this as a christmas gift when she was 2 .} it has been ready many times , and \mycolorbox{MistyRose1}{since been passed along to my son} who is now 4 . \mycolorbox{LightPink1}{my children enjoy} the tactile quality of the monkeys faces . \mycolorbox{IndianRed1}{it is helpful learning counting when there is something they can feel .} i have always enjoyed reading the sing song story. \mycolorbox{LightPink1}{it does not take long} to read , and after all these years \mycolorbox{LightPink1}{i pretty much} \mycolorbox{LightPink1}{have it memorized .}\mycolorbox{IndianRed1}{a great book , very fun .}
 \\ \\
        (3) & \mycolorbox{IndianRed1}{this mattress is too narrow to be comfortable . you fit on it fine but}\mycolorbox{MistyRose1}{ because of the air . i found} that it was a balancing act to switch positions . \mycolorbox{LightPink1}{i tried more and less air to no effect .}\mycolorbox{MistyRose1}{i think if you sleep} on your back and stay in that position \mycolorbox{IndianRed1}{it would be fine but unfortunately that is not} how i sleep . \mycolorbox{MistyRose1}{the strong} vinyl \mycolorbox{MistyRose1}{smell does go away after} airing out though.\\
        \hline
        \multicolumn{2}{c}{\multirow{2}{*}{(b) Summary view of the original text.}}\\ \\
	\end{tabular}
	\caption{Visualization of multi-view attention. Above is the heatmap of the sentiment-view attention, and below is the heatmap of the summary-view attention. Deeper colors means higher attention scores.}\label{visualization}
	\vspace{-0.1in}
\end{table*}

\section{Related Work}

Rush et al.~\shortcite{abs} first proposes an abstractive based summarization model, which uses an attentive CNN encoder to compress texts and a neural network language model to generate summaries.
Chopra et al.~\shortcite{ras} explores a recurrent structure for abstractive summarization. 
To deal with out-of-vocabulary problem, Nallapati et al.~\shortcite{ibmsummarization} proposes a generator-pointer model so that the decoder is able to generate words in source texts.
Gu et al.~\shortcite{copynet} also solves this issue by incorporating copying mechanism, allowing parts of the summaries to be copied from the source contents.
See et al.~\shortcite{See2017} further discusses this problem, and incorporates the pointer-generator model with the coverage mechanism.
Hu et al.~\shortcite{lcsts} builds a large corpus of Chinese social media short text summarization.
Chen et al.~\shortcite{distraction} introduces a distraction based neural model, which forces the attention mechanism to focus on the difference parts of the source inputs.
Ma et al.~\shortcite{MaEA2017} proposes a neural model to improve the semantic relevance between the source contents and the predicted summaries.

There are some work concerning with both summarization and sentiment classification. Hole and Takalikar~\shortcite{hole2013} and Mana et al.~\shortcite{mane2015} propose the models to produce both the summaries and the sentiment labels. However, these models train the summarization part and the sentiment classification part independently, and require rich hand-craft features.
Some work has improved the summarization with the help of classification.
Cao et al.~\shortcite{tcsum} proposes a model to train the summary generator and the text classifier jointly, but only improves the performance of the text summarization.
Titov and McDonald~\shortcite{Titov2008} proposes a sentiment summarization method to extract the summary from the texts given the sentiment class.
Lerman et al.~\shortcite{Lerman2009} builds a new summarizer by
training a ranking SVM model over the set
of human preference judgments, and improves the performance of sentiment summarization.
Different from all of these works, our model improves both text summarization and sentiment classification, and does not require any hand-craft features.

\section{Conclusions}

In this work, we propose a model to generate both the sentiment labels and the human-like summaries, hoping to summarize the opinions from the coarse-grained sentiment labels to the fine-grained word sequences. We evaluate our proposed model on several online reviews datasets. Experimental results show that our model achieves better performance than the baseline systems on both abstractive summarization and sentiment classification.

\section*{Acknowledgements}

This work was supported in part by National Natural Science Foundation of China (No. 61673028), National High Technology Research and Development Program of China (863 Program, No. 2015AA015404), and the National Thousand Young Talents Program. Xu Sun is the corresponding author of this paper.

\nocite{SunEA2017,SunWei2017,dnerre,amr,discourse,wean,unpair}

\bibliographystyle{named}
\bibliography{ijcai18}

\end{document}